\documentclass[letterpaper]{article} 
\usepackage{aaai24}  
\usepackage{times}  
\usepackage{helvet}  
\usepackage{courier}  
\usepackage[hyphens]{url}  
\usepackage{graphicx} 
\usepackage{natbib}  
\usepackage{caption} 
\usepackage{algorithm}
\usepackage{algorithmic}

\usepackage{graphicx}
\usepackage{multirow}
\usepackage{relsize}

\usepackage{newfloat}
\usepackage{listings}
\DeclareCaptionStyle{ruled}{labelfont=normalfont,labelsep=colon,strut=off} 
\floatstyle{ruled}
\newfloat{listing}{tb}{lst}{}
\floatname{listing}{Listing}
\title{Auto311: A Confidence-Guided Automated System for Non-emergency Calls}
\author {
    Zirong Chen,
    Xutong Sun,
    Yuanhe Li,
    Meiyi Ma
}
\affiliations {
    Department of Computer Science, Vanderbilt University, Nashville, Tennessee 37235, USA\\
    \{zirong.chen, xutong.sun, yuanhe.li, meiyi.ma\}@vanderbilt.edu
}

\usepackage{bibentry}

\begin{document}

\maketitle

\begin{abstract}
Emergency and non-emergency response systems are essential services provided by local governments and critical to
protecting lives, the environment, and property. The effective handling of (non-)emergency calls is critical for public safety and well-being. 
By reducing the
burden through non-emergency callers, residents in critical need of assistance through 911 will receive fast
and effective response. 
Collaborating with the Department of Emergency Communications (DEC) in Nashville, we analyzed 11,796 non-emergency call recordings and developed Auto311\footnote{Code and Demo: \url{https://github.com/AICPS-Lab/Auto311}}, the first
automated system to handle 311 non-emergency calls, which (1) effectively and dynamically predicts ongoing non-emergency incident types to generate tailored case reports during the call; (2) itemizes essential information from dialogue contexts to complete the generated reports; and (3) strategically structures system-caller dialogues with optimized confidence. We used real-world data to evaluate the system's effectiveness and deployability. The experimental results indicate that the system effectively predicts incident type with an average F-1 score of 92.54\%. Moreover, the system successfully itemizes critical information from relevant contexts to complete reports, evincing a 0.93 average consistency score compared to the ground truth. Additionally, emulations demonstrate that the system effectively decreases conversation turns as the utterance size gets more extensive and categorizes the ongoing call with 94.49\% mean accuracy. 

\end{abstract}
\section{Introduction}

Emergency and non-emergency response systems are essential services provided by local governments and critical to protecting lives, the environment, and property. While 911 is primarily used for emergency services, 311 is a non-emergency phone number that people can call to find information about municipal services, make complaints, or report problems like stolen property, road damage, etc. Both emergency and non-emergency calls are operated by the Department of Emergency Communication (DEC) in most cities. DECs across the nation receive an overwhelmingly high number of calls, with the national yearly average of 911 calls being close to 240 million~\cite{nycnextgen911, ma2019data}. The growing use of response systems comes at a time when local governments face increasing pressure to do more with fewer resources. Indeed, the number of local government employees in the United States has shrunk by nearly 5\% in 2021~\cite{2021annual}, and a large proportion of counties and municipalities anticipate a significant general fund shortfall as the United States transitions out of the COVID pandemic~\cite{g12afonso2021planning}.

To mitigate this issue, we introduce Auto311, the first automated system to handle 311 non-emergency calls. Auto311 features two key components: incident type detection and information itemization. It identifies probable incident types from ongoing calls and dynamically generates and updates case reports accordingly. Simultaneously, these reports direct the information itemization module to gather necessary information, streamlining the process.


Previous works have aimed to optimize (non-)emergency management \cite{sun2020applications, wex2014emergency, manoj2007communication, 
ma2021novel, 
ma2021toward, 
ma2018cityresolver, 
chen2008coordination}. However, most of those works focus more on the emergency resource allocation after one case report is placed, for example, route optimization for ambulances and response center localization for faster responses \cite{mukhopadhyay2022review}. Although other available could frameworks like AWS's Lex\footnote{AWS Lex: \url{https://aws.amazon.com/lex/}} and Google's DialogFlow\footnote{Diagflow: \url{https://cloud.google.com/dialogflow?hl=en}} can set up automated dialogue service within a few hours, they require clear and relatively fixed dialogue charts to guide the conversation. When it comes to non-emergency call handling, leaving alone privacy and safety issues brought by online solutions, most incident types have unique dialogue charts compared to others, making it unrealistic to separately handle the conversation for each incident type. However, Auto311, at a system level, takes full advantage of the emitted confidence scores of each component to strategically optimize the dialogue. 


However, developing Auto311 poses some key \textbf{challenges}. First, unlike plenty of work that has been done to solve text classification problems \cite{kowsari2019text, mironczuk2018recent, minaee2021deep, aggarwal2012survey} with outputting the most likely one category in the end, the incident prediction in this task has to cope with calls involving multiple incident types instead, see Section Motivating Study for more details. Second, although measuring confidence in machine learning models has become more and more popular recently, refer to Section Related Work for more details, there still lacks an effective method to measure the confidence score behind the model outputs in a textual format. Lastly, although pre-trained models yield satisfying performance on datasets with general purposes, e.g., Bert \cite{devlin2018bert} on SQuAD \cite{rajpurkar2016squad}, Auto311 has to align with more task-specific data and goals under the non-emergency call handling scope. We summarize our \textbf{contributions} as follows:
\begin{itemize}
    \item We annotate and analyze 11,796 authentic audio recordings of non-emergency calls.
    \item We build Auto311, the first confidence-guided automated system to handle non-emergency calls by navigating the conversation with optimized confidence, dynamically predicting incident type, and generating reports with key information. 
    
    
    
    \item We evaluate the performance of Auto311 using real-world recordings of non-emergency. It achieves an average F-1 score of 92.54\% on the incident type prediction and an average score of 0.93 for information itemization.
    \item We emulate the usage of Auto311 using recordings from our dataset and analyze Auto311's system-level impacts. Auto311 dynamically adjusts to shifting incident types, reduces follow-up conversations, and yields an overall average accuracy of 94.49\% for categorizing the emulated call utterances.
\end{itemize}

\section{Motivating Study} 


We annotate and analyze 11,796 real-world recordings of non-emergency calls, and make two important observations. 

\subsubsection{Unintentional Additional Information}

 Examination of audio transcriptions indicates callers tend to provide supplementary details beyond the dispatcher's specific inquiries. We found that 72\% of callers offer extra information, exceeding the question's scope in our dataset. Notably, this additional information can enhance the precision and comprehensiveness of the emergency response system. See the example below with the caller's personal information removed: 

$\mathsf{Dispatcher}$: Metro Nashville Police, Fire, and Medical, what is the location of your emergency? $\mathsf{Caller}$: Oh, I'm not sure if this is an emergency. I am \textit{\#name}, \textit{\#phone\_number}. The address is \textit{\#address}. It's the King Buffet. I saw a customer out in the parking lot smoking crackpipes in front of all the customers.


In this conversation turn, although the dispatcher only inquires about location, the caller provides additional details like name, phone number, and suspicious activity, suggesting a drug-related case. Strategically leveraging such extra information could optimize emergency response conversations by proactively addressing potential follow-up questions, thus streamlining the interaction. 


\subsubsection{Shifting and Multiple Incident Types}
Call recordings show that ongoing incidents mentioned by callers occasionally encompass multiple incident types (at a rate of 38.27\%). Callers also tend to modify incident types as they uncover more details. For instance, phrases like ``someone busted my car and my wallet is gone'' indicate two incident types – damaged and lost stolen property. Similarly, in ``I saw a car illegally parked... oh, wait, it's abandoned because the bumper is off and rusted,'' the caller initially reports illegal parking, then recognizes it as an abandoned vehicle based on new details. This underscores the need for our system to recognize various incident types concurrently and adapt to evolving conversations.

\section{Overview of Auto311}

\begin{figure}[h]
    \centering
    \includegraphics[width=0.45\textwidth]{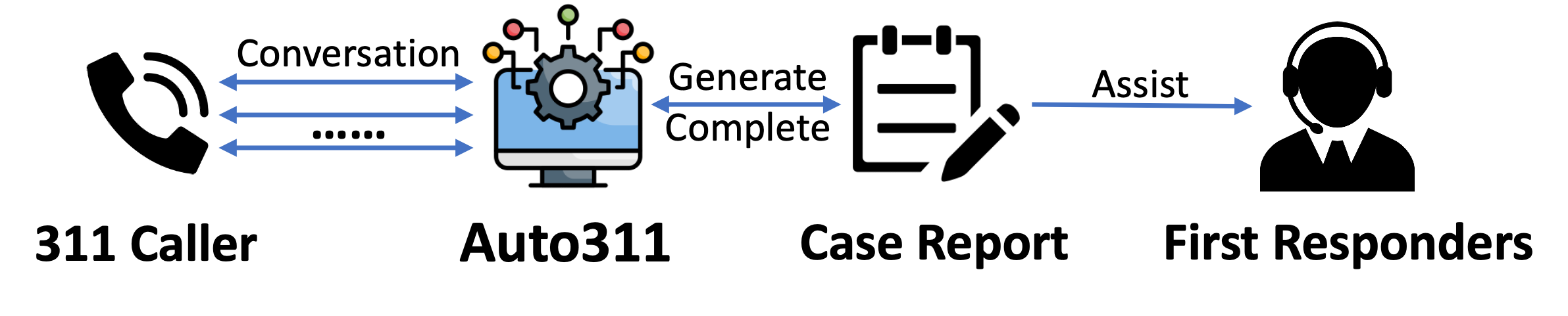}
    \caption{Auto311 in Emergency Response}
    \label{fig:sys_overview}
\end{figure}

Auto311 is designed to automatically handle non-emergency calls by engaging in interactive conversations with callers. An outline of our system's structure is shown in Figure \ref{fig:sys_overview}. The system comprises five key components: the \textit{conversational interface} for interacting with callers, the \textit{handover control} to transfer calls to human operators if needed, the \textit{incident type prediction} module to identify probable incident types, the \textit{information itemization} module to organize details in case reports, and the \textit{confidence-guided report generation and dialogue optimization} module for generating informed reports and optimizing subsequent conversations using confidence guidance.

\begin{figure*}[t]
    \centering
    \includegraphics[width=0.95\textwidth]{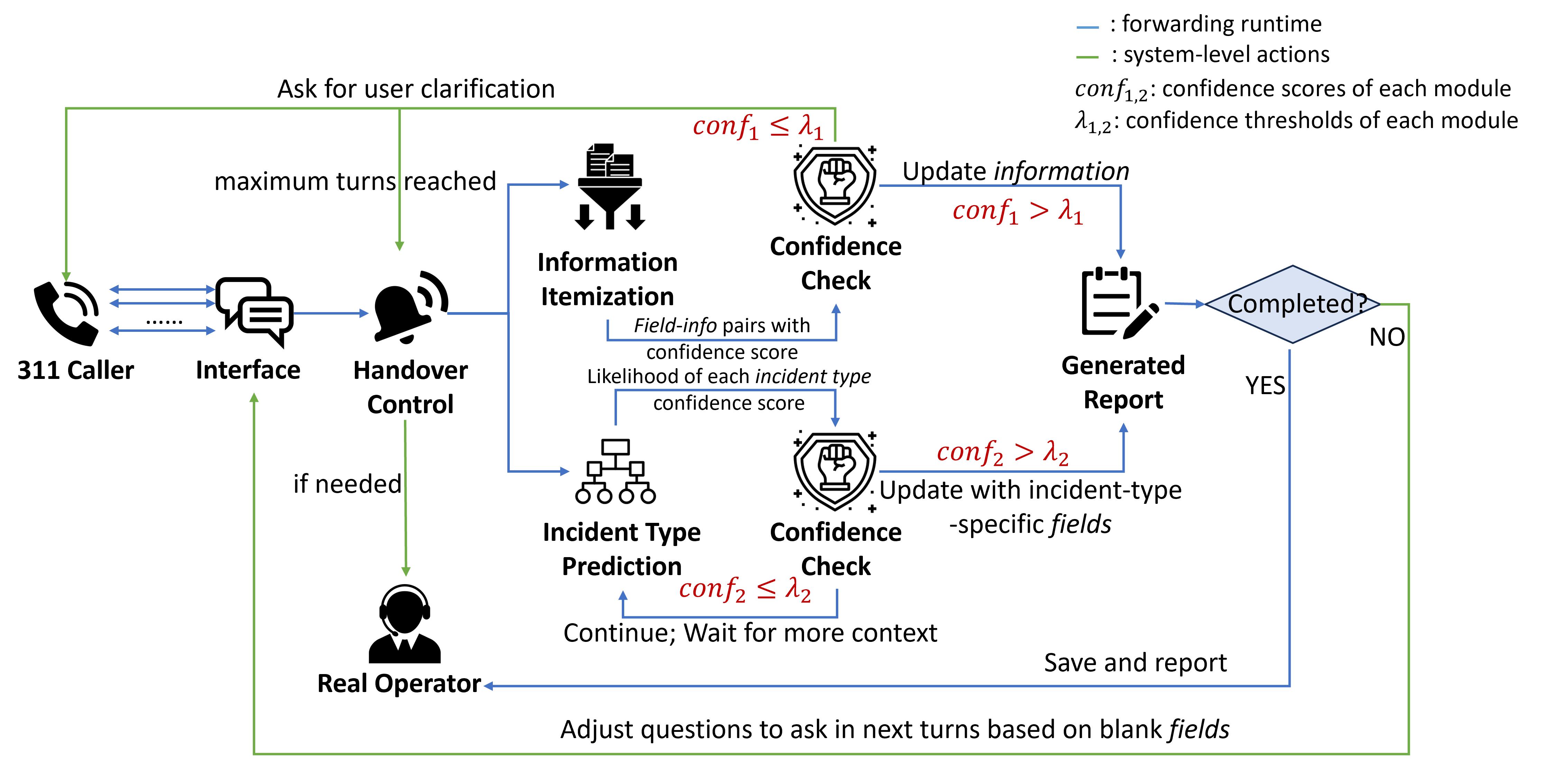}
    \caption{Confidence-guided System Design}
    \label{fig:sys_logic}
\end{figure*}

At runtime, as shown in Figure \ref{fig:sys_logic}, when a caller initiates contact, the \textit{conversational interface} starts the dialogue with opening questions, collecting essential information like the caller's name and incident location. Each response from the caller forms an utterance. Subsequently, the \textit{handover control} function evaluates whether human operator intervention is required for the call. The call proceeds to subsequent modules only if the handover is not needed. At the same time, the \textit{incident type prediction} module forecasts the most likely incident type(s) for report generation, while the \textit{information itemization} module provides potential details from the caller's utterance to fill the report's sections. Additionally, the \textit{confidence-guided report generation and dialogue optimization} module constantly monitors confidence scores to ensure the report is filled with high confidence, thereby optimizing follow-up conversations.

Importantly, the case report referred to here differs from the question list within the conversational interface. The question list stores upcoming inquiries for the interface, while the case report shapes the subsequent dialogue by updating the question list with its incomplete sections after each conversation turn.



\subsection{Conversational Interface}

The conversational interface supports both voice and text inputs. Employing state-of-the-art audio transcription tools, notably the OpenAI Whisper model \cite{openai_whisper_2022}, this interface adeptly transforms speech into text, accommodating a range of accents. For text-to-speech functionality, we harness advanced audio generation tools like the Suno-AI Bark model \cite{bark}, acclaimed for producing realistic voices within a lightweight model framework. Beyond speech-to-text and text-to-speech conversion, the interface engages in conversations using a dynamically updated question list. This list determines the query sequence and guides the interface's speech-to-text conversion process.

\subsection{Always-on Handover Control}

The handover control module remains active throughout runtime, redirecting calls to human operators when necessary. Collaborating with DEC, we identify specific scenarios that activate this module: (1) downstream module exceptions, like uncertain information; (2) caller's repeated request for human interaction; (3) proactive alerts for potential urgency. The first case is managed within system actions. For example, when Auto311 seeks clarification due to uncertain details, it limits such queries to three turns. Exceeding this threshold triggers exceptions and activates handover control. Addressing the other two cases, we develop an interpretable rule-based detection mechanism, prioritizing interpretability and control. Using Latent Dirichlet Allocation (LDA) \cite{blei2003latent}, we curate a sensitive word list through manual review. Our approach combines natural language processing (NLP) features \cite{nltk}, such as stemming, lemmatization, part-of-speech tags, and shallow parsing, with custom patterns to establish rules activating handover control, thus ending system interaction. More details are available in the appendix. Note that patterns and sensitive words are not exhaustive, allowing future expansion of trigger conditions. However, the broader process is beyond this paper's scope.


\subsection{Incident Type Prediction}
\label{subsec:prediction}

The incident type prediction module utilizes contextual information from previous caller utterances. The module takes the overall context, covering all prior utterances as input. Since a context can involve multiple incident types, we apply a multi-layer hierarchical structure and bootstrap-like procedure for classification. This tracks the possibility of the call belonging to each incident type (see Confidence-guided System Design for details). The hierarchical structure and iterative procedure enable the prediction module to handle multiple incident types per call using full conversational context.


\subsection{Information Itemization}
\label{subsec:itemization}

The information itemization module completes empty case report sections by quoting the caller's utterances. This involves narrative fields seeking explanatory details and yes/no fields confirming facts. For narratives, we leverage extractive question-answering frameworks - the blank fields are inputs, and outputs quote relevant caller utterances. Yes/no fields become binary classification, predicting yes or no from the last utterance. Unlike incident prediction using all contexts, itemization considers only the latest utterance. 



\subsection{Confidence-guided Report Generation and Dialogue Optimization}

This module updates the report and optimizes dialogues as the conversation progresses. See technical details in Section Confidence-guided System Design.

\section{Confidence-guided System Design}

This section delves into the technical aspects of confidence guidance within Auto311. Firstly, we explain the method to derive confidence scores from the machine learning models. Secondly, we elucidate the purpose of the generated confidence scores within the workflow.


\begin{figure*}[ht]
    \centering
    \includegraphics[width=0.94\textwidth]{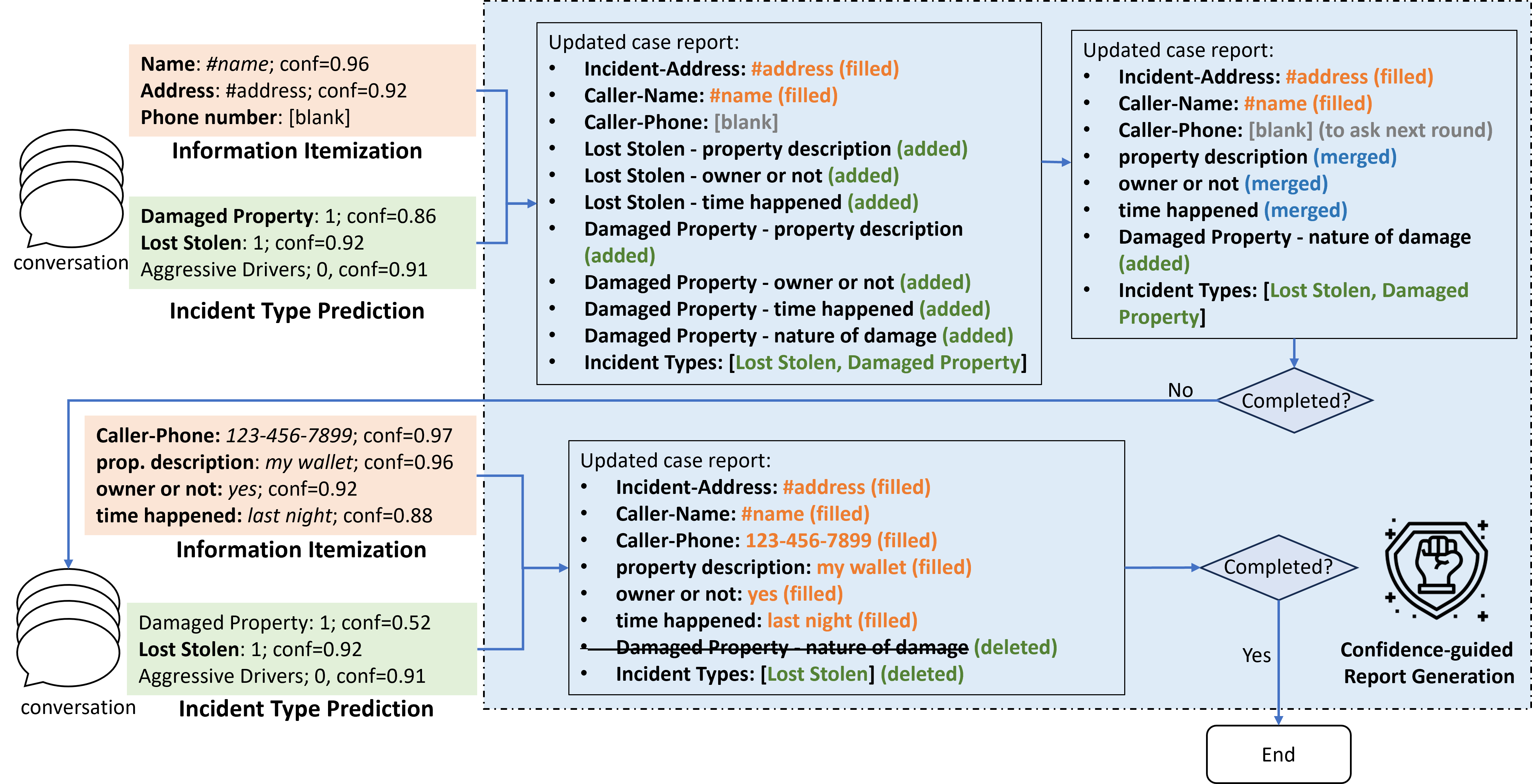}
    \caption{A General Case Study in Confidence Guidance}
    \label{fig:case_study}
\end{figure*}

\subsection{Confidence Measurement}

We define confidence as consistency over multiple trials with the same inputs. We leverage Monte Carlo Dropout\cite{gal2016dropout, ma2021predictive} to generate the confidence prediction. Specifically, dropout was set as active at test time and assesses consistency across trials to obtain scores. Preset thresholds determine if outputs are confident - meeting or exceeding the threshold means confident. Auto311 applies machine learning models to handle two major tasks component-wisely: incident type prediction and information itemization. Here we detail Auto311's approach to measuring confidence.


\subsubsection{Confidence in Incident Type Prediction}

To address potential multiple incident types within a call, we employ a multi-layer hierarchy structure coupled with a bootstrap-like process. The initial layer of this structure involves training a neural network to assess if the present context corresponds to the most common incident type. Subsequently, the second layer trains another neural network to determine if the context aligns with the second most frequent incident type, excluding the previously identified type. This pattern continues for subsequent types. As a result, (1) the structure can identify all possible incident types within a call; (2) each category operates independently, facilitating future adjustments based on new data or expansion to more categories. At runtime, we establish confidence scores by measuring consistency across output distributions for the same input, utilizing active dropout to quantify prediction uncertainty. The hierarchical cascade structure adeptly handles multiple types, while confidence scoring measures prediction certainty.


\subsubsection{Confidence in Information Itemization}


Regarding textual outputs for information itemization, determining confidence necessitates a consistency assessment between texts. Traditional text comparison methods often prioritize aspects like edit distances and length. See details in Related Work. However, our collaboration with DEC underscores the value of succinct outputs with ample details for case reports. Consider the scenario where the module generates ``on the 2525 West End Ave'' while the ground truth is ``2525 West End Ave.'' Traditional methods yield low scores, such as 0.5 from BLEU-bigram. However, when dispatchers gather incident location details, these outputs should exhibit high consistency due to matching location keywords and similar semantics. To address this, we adopt a new approach. For keywords, we employ an unsupervised state-of-the-art keyword extractor \cite{campos2018yake} to extract key segments from model outputs. The overlap between keyword segment lists is calculated. For semantics, we utilize SentenceBERT \cite{reimers-2019-sentence-bert} to project each output into a latent space, assessing the similarity between represented textual string lists. The overall score, calculated via Polyak averaging (p=0.2), integrates keyword overlaps and embedding distances. This metric enables us to gauge consistency and generate a confidence score.


\subsection{Report Generation and Dialogue Optimization}

Auto311 is designed to be confidence-guided. With maximum confidence guaranteed, it (1) dynamically updates case reports at every turn of the conversation and (2) guides the follow-up conversation based on the generated case report. See Figure \ref{fig:sys_logic} for more detailed system logic. 

Confidence drives precise report completion via information itemization. Using the latest caller utterance, it populates report details. Textual output confidence is determined through text comparison. As in Figure \ref{fig:sys_logic}, uncertainty ($conf_1 \leq \lambda_1$) prompts Auto311 to seek clarification for guiding questions, capped at three turns before handover control. Confidence ($conf_1 > \lambda_1$) skips further dialogue optimization for filled fields. Confidence also aids the incident type prediction module's adaptability and report efficiency. Using complete utterance context, it predicts likely incident types. As shown in Figure \ref{fig:sys_logic}, low confidence ($conf_2 \leq \lambda_2$) excludes uncertain predictions from reports. High confidence ($conf_2 > \lambda_2$) incorporates predictions. Systemically, prediction persists each turn, even post early confident identification. This iterative process tracks trends in types and scores, updating reports on confidence drops ($conf_2 \leq \lambda_2$). Confidence-driven adaptation detects and responds to evolving incident types during calls, further updating reports.

Previous component confidence scores further optimize future dialogues. In the case study illustrated in Figure \ref{fig:case_study}, when a new caller utterance is received, Auto311 identifies fields to complete in the case report (e.g., incident-address, caller-name, caller-phone). High-confidence completion marks them as done. Unfinished fields guide subsequent questions (e.g., requesting a callback number if caller-phone is missing). Concurrently, incident type is confirmed if confidence surpasses a set threshold (0.85 in Figure \ref{fig:case_study}). With the type established, specialized fields are identified (e.g., property description for lost/stolen cases), updating the report. Auto311 then prioritizes shared general fields (e.g., property description, time, ownership status), streamlining dialogue to focus initially on universal details. This avoids asking about the damage nature before finalizing the incident type, which applies only to damaged property cases. This optimization confirms type(s) with more context. In the example, the next turn's details indicate a shift to "lost/stolen" only. The report is updated accordingly. The dialogue concludes when all report fields are complete. Confidence scoring thus optimizes the flow by collecting universal details first and adapting to emergent incident types.

\section{Evaluation}

\begin{table*}[]
\centering
\small
\begin{tabular}{||c|c|c|c|c|c|c|c|c|c||}
\hline
                 & \begin{tabular}[c]{@{}c@{}}Minor\\ Crash\end{tabular} & \begin{tabular}[c]{@{}c@{}}Lost\\ Stolen\end{tabular} & \begin{tabular}[c]{@{}c@{}}Aggressive\\ Drivers\end{tabular} & \begin{tabular}[c]{@{}c@{}}Check\\ Welfare\end{tabular} & \begin{tabular}[c]{@{}c@{}}Damaged\\ Property\end{tabular} & \begin{tabular}[c]{@{}c@{}}Noise\\ Violation\end{tabular} & \begin{tabular}[c]{@{}c@{}}Roadway\\ Hazard\end{tabular} & \begin{tabular}[c]{@{}c@{}}Abandoned\\ Vehicles\end{tabular} & \begin{tabular}[c]{@{}c@{}}Drug-Pros\\ Activity\end{tabular} \\ \hline\hline
LSTM             & 56.47\%                                               & 0.00\%                                                & 53.85\%                                                & 82.35\%                                                 & 0.00\%                                                     & 0.00\%                                                    & 63.83\%                                                  & 69.23\%                                                      & 46.15\%                                             \\ \hline
CNN              & 75.86\%                                               & 85.71\%                                               & 72.72\%                                                & 81.08\%                                                 & 40.00\%                                                    & 44.44\%                                                   & 80.00\%                                                  & 62.07\%                                                      & 25.00\%                                             \\ \hline
RCNN             & 90.57\%                                               & 82.35\%                                               & 61.54\%                                                & 82.35\%                                                 & 76.92\%                                                    & 83.33\%                                                   & 86.67\%                                                  & 26.09\%                                                      & 25.00\%                                             \\ \hline
RNN              & 63.33\%                                               & 40.00\%                                               & 52.71\%                                                & 73.33\%                                                 & 44.44\%                                                    & 83.33\%                                                   & 74.29\%                                                  & 28.57\%                                                      & 26.67\%                                             \\ \hline
Self-Attn        & 88.46\%                                               & 88.89\%                                               & 66.67\%                                                & 87.50\%                                                 & 66.67\%                                                    & 50.00\%                                                   & 81.25\%                                                  & 9.52\%                                                       & 35.29\%                                             \\ \hline
Attention        & 91.69\%                                               & 62.50\%                                               & 50.00\%                                                & 53.66\%                                                 & 54.55\%                                                    & 60.00\%                                                   & 81.08\%                                                  & 66.67\%                                                      & 62.50\%                                             \\ \hline
Bert             & 95.04\%                                               & 91.50\%                                               & 92.31\%                                                & 90.00\%                                                 & 88.89\%                                                    & 83.33\%                                                   & 90.91\%                                                  & 94.12\%                                                      & 92.40\%                                             \\ \hline
\textbf{Auto311} & \textbf{95.71\%}                                      & \textbf{93.00\%}                                      & \textbf{93.75\%}                                       & \textbf{90.00\%}                                        & \textbf{94.12\%}                                           & \textbf{83.33\%}                                          & \textbf{90.91\%}                                         & \textbf{94.12\%}                                             & \textbf{92.40\%}                                    \\ \hline
\end{tabular}
\caption{Averaged performance (F-1) over 30 trials on incident type prediction}
\label{tab:classification}
\end{table*}

Experiments assess the performance of (1) confidence-guided incident prediction, (2) confidence-guided itemization, and (3) the overall system. Our dataset includes 11,796 non-emergency calls from the DEC in Nashville, TN. Metrics for incident type prediction are precision, recall, F-1, and accuracy. The newly introduced text comparison method evaluates itemization module outputs. The experiments were run on a machine with 2.50GHz CPU, 32GB memory, and Nvidia GeForce RTX 3080Ti GPU.



\subsection{Confidence-guided Incident Type Prediction}

This section aims to evaluate Auto311's performance on incident type prediction. Baselines use various neural networks like LSTM, CNN, RCNN, Self-Attention, Bahdanau's Attention, and BERT \cite{hochreiter1997long, kim2014convolutional, lai2015recurrent, vaswani2017attention, bahdanau2014neural, wolf2019huggingface}(see Table \ref{tab:classification}). We include 9 categories due to the page limit (full results including standard deviation stats in Appendix). Experiments comprehensively assess prediction with different model architectures on real call data.


Analysis shows traditional models like CNN perform poorly, with just 62.99\% average F-1 on these 9 types. Transcription diversity from varied callers increases task complexity (e.g., different speaking habits), challenging learning without prior knowledge. BERT surpasses other models, with 92.54\% average F-1 and 100\% max across all 11 types. Leveraging BERT's pre-trained weights and confidence guidance, Auto311 further improves BERT F-1 from 91.50\% to 93.00\% for lost/stolen cases. Results demonstrate prediction difficulties due to call diversity and Auto311's enhancements over BERT using confidence scoring.

\textit{In summary, based on the results, Auto311 effectively dispatches the ongoing call to the given incident types. In terms of F-1 score, the BERT backend has the most competitive results. Confidence guidance further improves performance.}

\subsection{Confidence-guided Information Itemization}

This experimental setup assesses the performance of the information itemization module of Auto311 using various backends (DistilBERT, BERT, RoBERTa, LongFormer, BigBird \cite{sanh2019distilbert, beltagy2020longformer, zaheer2020big}) and benchmarks (SQuAD, CUAD, TriviaQA \cite{rajpurkar2016squad, rajpurkar2018know, hendrycks2021cuad, joshi2017triviaqa}) and compares it to large language models (LLMs) like GPT3.5 and 4\footnote{Prompt: ``\textit{I will provide context and a set of questions. Please respond to the questions using exact quotes from the context. Your answers should be concise and comprehensive. Context: ...; Question Set: ...''}}, with results in Table \ref{tab:informationextraction}. We utilize our consistency score in this evaluation. Two types of data samples are evaluated for performance: random test samples from archived data (collected by the end of 2022) and the latest data samples (collected from the beginning of 2023) from the call center. Recognizing that city-related information evolves (e.g., new places, activities), we simulate Auto311's usage under knowledge evolution by assessing performance on the latest data samples. Furthermore, we assess Auto311's performance in information itemization when queried with various fields, encompassing basic fields, less specific fields, and more specific fields. Basic fields include essential data like incident location, less specific fields cover descriptors like vehicle and human/suspect descriptions, and more specific fields pertain to incident-specific details such as the timing of the incident (DamagedProperty-when).

\begin{table}[h]
\centering
\small
\begin{tabular}{||c|c|c||}
\hline
           & Archived (test) & Latest (runtime)  \\ \hline\hline
DistilBERT-SQuAD2 &  0.5546    & 0.5330                        \\ \hline        
BERT-SQuAD2 &   0.2422   & 0.2791                        \\ \hline
RoBERTa-SQuAD2 &   0.1172   & 0.2581                        \\ \hline
RoBERTa-CUAD &   0.2188   & 0.2378                        \\ \hline
LongFormer-TriviaQA &   0.3260   & 0.1424                        \\ \hline
BigBird-TriviaQA &   0.5289   &     0.5015                    \\ \hline
GPT3.5 (June 2023)     &   0.6343   & 0.6529                         \\ \hline
GPT4 (June 2023)      &   0.6578   & 0.6264                          \\ \hline
\textbf{Auto311}   &   \textbf{0.9329}   & \textbf{0.8605}          \\ \hline
\end{tabular}
\caption{Performance on information itemization}
\label{tab:informationextraction}
\end{table}

\begin{table*}[t]
\centering
\small
\begin{tabular}{||c|ccccccccc||}
\hline
                  & \multicolumn{3}{c|}{Basic Fields}                                                                                                                                                                                                                                                                                           & \multicolumn{6}{c||}{Less Specific Feilds}                                                                                                                                                                                  \\ \hline\hline
                  & \multicolumn{1}{c|}{Inc-Loc}                                                                & \multicolumn{1}{c|}{Caller-Name}                                                                           & \multicolumn{1}{c|}{Caller-Phone}                                                                   & \multicolumn{2}{c|}{Veh Desc}                                       & \multicolumn{2}{c|}{Human/Suspect Desc}                                 & \multicolumn{2}{c||}{Prop Desc}                \\ \hline
w/o Conf Guide    & \multicolumn{1}{c|}{0.9035}                                                                           & \multicolumn{1}{c|}{0.9478}                                                                                & \multicolumn{1}{c|}{1.0000}                                                                            & \multicolumn{2}{c|}{0.8538}                                                    & \multicolumn{2}{c|}{0.9678}                                                    & \multicolumn{2}{c||}{0.9512}                              \\ \hline
w/ Conf Guide     & \multicolumn{1}{c|}{\textbf{0.9631}}                                                                  & \multicolumn{1}{c|}{\textbf{0.9962}}                                                                       & \multicolumn{1}{c|}{\textbf{1.0000}}                                                                   & \multicolumn{2}{c|}{\textbf{0.9104}}                                           & \multicolumn{2}{c|}{\textbf{1.0000}}                                           & \multicolumn{2}{c||}{\textbf{1.0000}}                     \\ \hline\hline
                  & \multicolumn{9}{c|}{More Specific Fields}                                                                                                                                                                                                                                                                                                                                                                                                                                                                                                                \\ \hline\hline
\multirow{2}{*}{} & \multicolumn{1}{c|}{\multirow{2}{*}{\begin{tabular}[c]{@{}c@{}}DamgProp\\ -When\end{tabular}}} & \multicolumn{1}{c|}{\multirow{2}{*}{\begin{tabular}[c]{@{}c@{}}AggDriver\\ -Behavior\end{tabular}}} & \multicolumn{1}{c|}{\multirow{2}{*}{\begin{tabular}[c]{@{}c@{}}CheckWel\\ -Relation\end{tabular}}} & \multicolumn{3}{c|}{\begin{tabular}[c]{@{}c@{}}MinorCrash\\ -BlockTraffic (Y/N)\end{tabular}}                          & \multicolumn{3}{c||}{\begin{tabular}[c]{@{}c@{}}CheckWel\\ -InpersonMeet (Y/N)\end{tabular}}   \\ \cline{5-10} 
                  & \multicolumn{1}{c|}{}                                                                                 & \multicolumn{1}{c|}{}                                                                                      & \multicolumn{1}{c|}{}                                                                                  & \multicolumn{1}{c|}{P}                 & \multicolumn{1}{c|}{R}                & \multicolumn{1}{c|}{F-1}              & \multicolumn{1}{c|}{P}                 & \multicolumn{1}{c|}{R}                & F-1              \\ \hline
w/o Conf Guide    & \multicolumn{1}{c|}{0.9045}                                                                           & \multicolumn{1}{c|}{0.8255}                                                                                & \multicolumn{1}{c|}{0.9023}                                                                            & \multicolumn{1}{c|}{66.67\%}          & \multicolumn{1}{c|}{100.00\%}          & \multicolumn{1}{c|}{80.0\%}           & \multicolumn{1}{c|}{83.33\%}           & \multicolumn{1}{c|}{83.33\%}          & 83.33\%          \\ \hline
w/ Conf Guide     & \multicolumn{1}{c|}{\textbf{1.000}}                                                                   & \multicolumn{1}{c|}{\textbf{0.9164}}                                                                       & \multicolumn{1}{c|}{\textbf{0.9855}}                                                                   & \multicolumn{1}{c|}{\textbf{88.89\%}} & \multicolumn{1}{c|}{\textbf{100.00\%}} & \multicolumn{1}{c|}{\textbf{94.12\%}} & \multicolumn{1}{c|}{\textbf{83.33\%}} & \multicolumn{1}{c|}{\textbf{100.00\%}} & \textbf{90.91\%} \\ \hline
\end{tabular}
\caption{Auo311's performance on different fields}
\label{tab:confidence_on_different_question_types}
\end{table*}


Table \ref{tab:informationextraction} yields the following insights: (1) pretrained model limitations: DistilBERT and BERT struggle in current non-emergency dispatch scenarios, performing notably lower than other methods. For instance, BERT pretrained on SQuAD achieves only 0.2422 on the test batch; (2) LLMs vs. Auto311: Despite LLMs' general NLP success, Auto311 consistently outperforms them on both datasets. For example, Auto311's performance surpasses GPT3.5 by 47\% on archived samples and 41\% on the latest samples; (3) adaptation to evolving knowledge: Auto311's 37\% performance lead over GPT4, despite a minor drop, underscores its proficiency in capturing evolving local city knowledge.


Table \ref{tab:confidence_on_different_question_types} highlights how confidence guidance enhances itemization across field types, e.g., improving consistency from 0.8255 to 0.9164 for aggressive driver behavior details. Auto311 achieves 100\% recall for binary questions, correctly predicting traffic blockage and in-person meetup needs. In real-world non-emergency scenarios, prioritizing high recall ensures comprehensive coverage of potential requests. Results showcase confidence scoring's role in optimizing itemization, particularly for critical binary fields.



\textit{In summary, the results underscore the difficulty of the information itemization task for both pretrained models and existing LLMs. Auto311's fine-tuning on our dataset effectively integrates task-specific knowledge, leading to competitive performance surpassing LLMs. Moreover, through runtime emulation, Auto311 adapts to evolving city knowledge, indicating potential for long-term deployment. Additionally, confidence guidance empowers Auto311 to enhance the completion of various case report fields.}

\subsection{System Level Performance}

The subsequent experiments focus on assessing Auto311's system-level performance. Due to data limitations, audio recordings only capture fixed dialogue paths, preventing direct interaction with the same caller in identical call scenarios. Instead, we emulate conversations by merging utterance segments. This emulation facilitates evaluating Auto311's capabilities in two aspects: (1) assessing its management of changing incident types and (2) analyzing the optimization of follow-up dialogues during emulation.

\subsubsection{Adjustments to Shifting Incident Types}

\begin{figure}[h]
    \centering
    \includegraphics[width=0.45\textwidth]{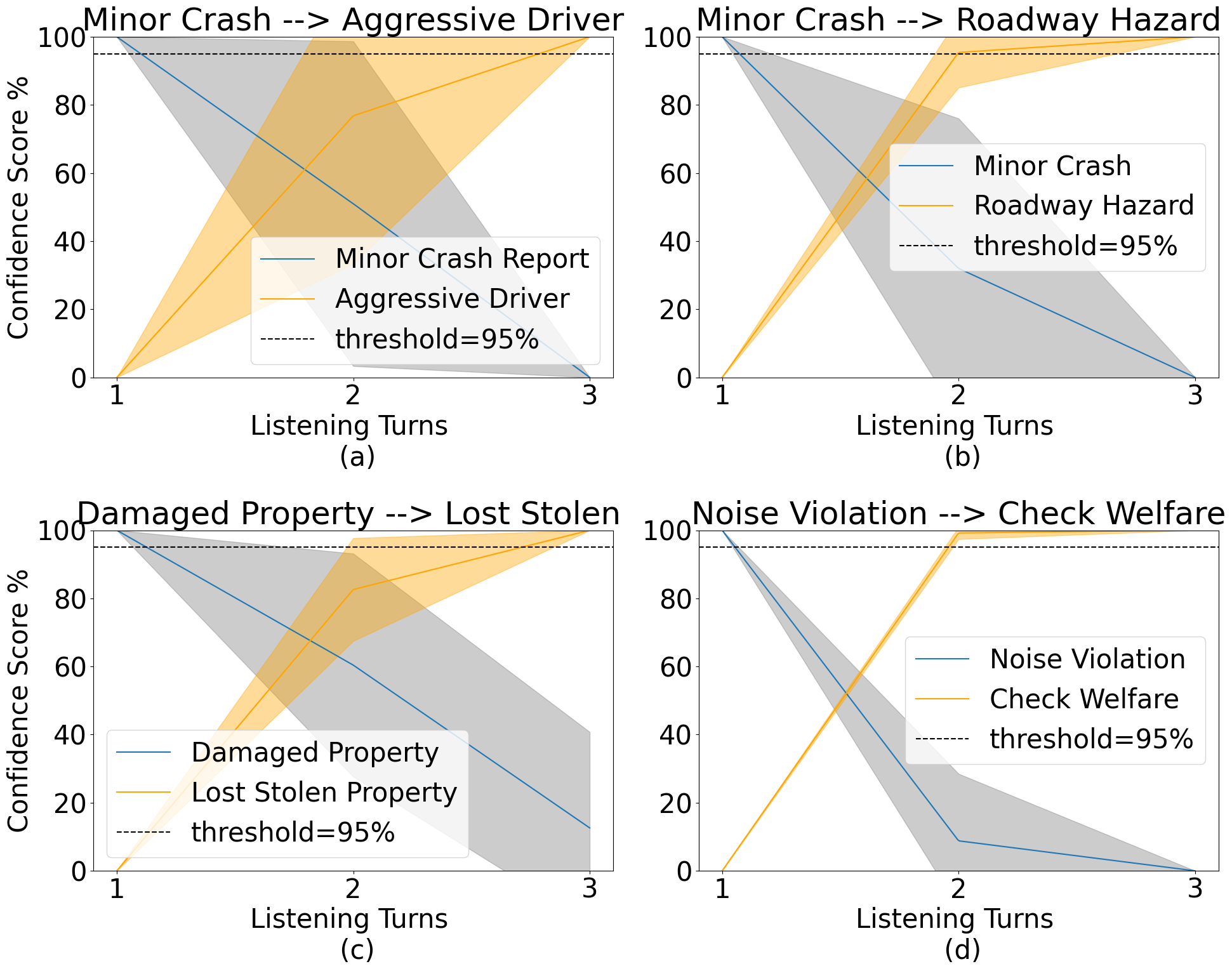}
    \caption{Confidence Changes in Shifting Incident Types}
    \label{fig:conf_curve}
\end{figure}

We evaluate Auto311's adaptability to shifting incident types. Using common shifts observed in the dataset (see Figure \ref{fig:conf_curve}), we emulate conversations where the caller firmly states type A (blue lines and grey regions), then adds type-specific details for type B (orange lines and regions), indicating a type shift. These experiments assess Auto311's real-time adaptation to emergent types in emulated interactions.

From Figure \ref{fig:conf_curve}, we observe that first, Auto311 handles all four major shifting situations in three future turns with more type-specific descriptions being fed as input to the incident type prediction module. Second, with the introduction of confidence guidance, Auto311 adjusts the prediction results to align with the shifting trend dynamically.

\textit{In summary, Auto311 adeptly handles shifting incident types in simulations, updating its understanding and creating optimized reports with more specific information over 3 follow-up turns.}


\subsubsection{Optimizations to Upcoming Dialogues}

\begin{figure}[h]
    \centering
    \includegraphics[width=0.42\textwidth]{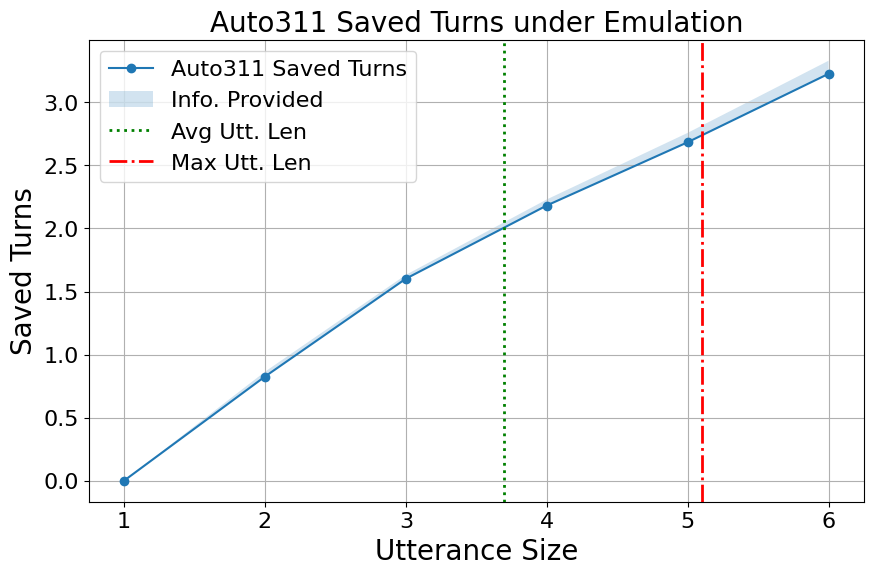}
    \caption{Emulated Usage of Auto311}
    \label{fig:emulation}
\end{figure}

We emulate Auto311 usage by composing caller utterances and assessing the relationship between saved turns and utterance size (see Figure \ref{fig:emulation}). Utterance size represents the count of past segments included. Across 100 emulations per size, we monitor saved turns and categorization accuracy. These experiments evaluate Auto311's ability to optimize dialogues through accumulated context and to make type predictions effectively.



Longer composed utterances contain more itemizable details. The blue line indicates Auto311's saved turns during emulation, while the light blue region represents total information provided. The average and maximum real-world utterance lengths are denoted by green and red dashed lines. Emulation demonstrates Auto311 effectively using additional information in caller utterances to minimize follow-up turns. Furthermore, Auto311 achieves a 94.49\% accuracy (not shown in Figure \ref{fig:emulation}) when handling composed utterances.



\textit{In summary, these emulations show our solution, at a system level, not only piratically optimizes future conversations by utilizing additional information provided in caller utterances but also effectively categorizes the potential incident types with an overall accuracy of 94.49\%.}

\section{Related Work}

\textbf{Question Answering and Large Language Models.} In recent years, advanced question-answering systems have evolved across various scenarios \cite{chen2023cityspecshield, chen2022intelligent, chen2022cityspec, diefenbach2018core}. Black-box abstractive QA systems like mBART and T5 \cite{chipman2022mbart, raffel2019t5} lack output control. Although large language models, like Claude\footnote{Anthropic Claude: \url{https://claude.ai}}, especially for QA dialogues \cite{brown2020language, ouyang2022training}, gain attention, we still argue that prompt-based models are unsuitable for emergency response due to compromised input preservation and advocate for a transparent, controllable offline approach prioritizing reliability and decision transparency.

\noindent\textbf{Confidence Scores in Machine Learning}. 
While significant efforts have been dedicated to assessing the model confidence \cite{poggi2017quantitative, hullermeier2021aleatoric, poggi2016learning}, we redefine confidence as internal consistency across identical inputs, deviating from common definitions. For text classification like call dispatching, this consistency is seen in distributional shifts. However, quantifying and analyzing output text changes across domains remains challenging in current information itemization setups. Most open-source QA models provide confidence scores for single runs, like HuggingFace's Bert-QA \cite{huggingface_bert_qa_2022}, which measures confidence in a single trial through simple multiplication of softmax distributions. Hence, a robust confidence measurement mechanism for Auto311 in incident type prediction and information itemization is crucial.

\noindent\textbf{Metrics for Text Comparison}. Many text comparison metrics are unsuitable for Auto311's information itemization. For our goal of concise, detailed outputs that allow deviations from the ground truth, metrics like Damerau-Levenshtein distance \cite{damerau1964technique} and BLEU \cite{papineni2002bleu} fall short. N-gram metrics like ROUGE \cite{lin-2004-rouge} and WER lack semantic understanding. Although end-to-end metrics like embeddings and learned metrics \cite{reimers-2019-sentence-bert, cer2018universal, artetxe-etal-2019-laser} consider semantics, they misalign with our criteria and lack interpretability and generalization in emergency response. Thus, a metric that gauges key information coverage from user utterances while meeting dispatch center requirements becomes essential.

\section{Summary}

In this paper, we introduce Auto311, the first automated system tailored for non-emergency call management. Our evaluations with real-world and emulated interactions show strong performance in (1) incident type prediction, (2) case report generation, and (3) enhanced follow-up conversations using confidence-based guidance. 
In future work, we will enhance Auto311 and deploy it to handle non-emergency calls in the real world.
By reducing the 
burden through non-emergency callers, residents in critical need of assistance through 911 will receive a fast and effective response.

\section{Acknowledgements}

This material is based upon work supported by the National Science Foundation (NSF) under Award Numbers 2228607. 
This work is a collaborative effort, and we are grateful for the support and contributions of everyone involved. In particular, we would like to express our gratitude for the valuable input and expertise provided by Stephen Martini, Director of the Department of Emergency Communication, and his team throughout the project. We also sincerely appreciate Keith Durbin, Chief Information Officer and Director of Information Technology Services, and Colleen Herndon, Assistant Director of GIS \& Data Insights, Information Technology Services for the Metropolitan Government of Nashville and Davidson County, for their valuable collaboration and insights.

\bibliography{auto311}

\begin{thebibliography}{56}
\providecommand{\natexlab}[1]{#1}

\bibitem[{Afonso(2021)}]{g12afonso2021planning}
Afonso, W. 2021.
\newblock Planning for the unknown: Local government strategies from the fiscal year 2021 budget season in response to the COVID-19 pandemic.
\newblock \emph{State and Local Government Review}, 53(2): 159--171.

\bibitem[{Aggarwal and Zhai(2012)}]{aggarwal2012survey}
Aggarwal, C.~C.; and Zhai, C. 2012.
\newblock A survey of text classification algorithms.
\newblock \emph{Mining text data}, 163--222.

\bibitem[{Artetxe et~al.(2019)Artetxe, Schwenk, Marquez, and Cho}]{artetxe-etal-2019-laser}
Artetxe, M.; Schwenk, H.; Marquez, L.; and Cho, K. 2019.
\newblock Massively Multilingual Sentence Embeddings for Zero-Shot Cross-Lingual Transfer and Beyond.
\newblock In \emph{Proceedings of the 2019 Conference on Empirical Methods in Natural Language Processing and the 9th International Joint Conference on Natural Language Processing (EMNLP-IJCNLP)}, 3293--3302.

\bibitem[{Bahdanau, Cho, and Bengio(2014)}]{bahdanau2014neural}
Bahdanau, D.; Cho, K.; and Bengio, Y. 2014.
\newblock Neural machine translation by jointly learning to align and translate.
\newblock \emph{arXiv preprint arXiv:1409.0473}.

\bibitem[{Beltagy, Peters, and Cohan(2020)}]{beltagy2020longformer}
Beltagy, I.; Peters, M.~E.; and Cohan, A. 2020.
\newblock Longformer: The long-document transformer.
\newblock \emph{arXiv preprint arXiv:2004.05150}.

\bibitem[{Bird, Loper, and Klein(2009)}]{nltk}
Bird, S.; Loper, E.; and Klein, E. 2009.
\newblock {NLTK}: The Natural Language Toolkit.
\newblock \url{http://www.nltk.org/}.
\newblock Online; accessed [insert date].

\bibitem[{Blei, Ng, and Jordan(2003)}]{blei2003latent}
Blei, D.~M.; Ng, A.~Y.; and Jordan, M.~I. 2003.
\newblock Latent Dirichlet Allocation.
\newblock \emph{Journal of Machine Learning Research}, 3: 993--1022.

\bibitem[{Bredin(2020)}]{Bredin_pyannote_2020}
Bredin, H. 2020.
\newblock pyannote.speaker: Speaker Diarization.
\newblock Accessed: 2023-08-03.

\bibitem[{Brown et~al.(2020)Brown, Mann, Ryder, Subbiah, Kaplan, Dhariwal, Neelakantan, Shyam, Sastry, Askell et~al.}]{brown2020language}
Brown, T.; Mann, B.; Ryder, N.; Subbiah, M.; Kaplan, J.~D.; Dhariwal, P.; Neelakantan, A.; Shyam, P.; Sastry, G.; Askell, A.; et~al. 2020.
\newblock Language models are few-shot learners.
\newblock \emph{Advances in neural information processing systems}, 33: 1877--1901.

\bibitem[{Campos et~al.(2018)Campos, Mangaravite, Pasquali, Jorge, Nunes, and Jatowt}]{campos2018yake}
Campos, R.; Mangaravite, V.; Pasquali, A.; Jorge, A.~M.; Nunes, C.; and Jatowt, A. 2018.
\newblock YAKE! Collection-Independent Automatic Keyword Extractor.
\newblock In \emph{European Conference on Information Retrieval}. Springer.

\bibitem[{Cer et~al.(2018)Cer, Yang, Kong, Hua, Limtiaco, John, Ngu, Christopher, Constant, Guajardo-Cespedes et~al.}]{cer2018universal}
Cer, D.; Yang, Y.; Kong, S.-y.; Hua, N.; Limtiaco, N.; John, R.; Ngu, A.; Christopher, A.; Constant, N.; Guajardo-Cespedes, M.; et~al. 2018.
\newblock Universal Sentence Encoder.
\newblock In \emph{Proceedings of the 2018 Conference on Empirical Methods in Natural Language Processing: System Demonstrations}, 169--174.

\bibitem[{Chen et~al.(2008)Chen, Sharman, Rao, and Upadhyaya}]{chen2008coordination}
Chen, R.; Sharman, R.; Rao, H.~R.; and Upadhyaya, S.~J. 2008.
\newblock Coordination in emergency response management.
\newblock \emph{Communications of the ACM}, 51(5): 66--73.

\bibitem[{Chen et~al.(2022{\natexlab{a}})Chen, Li, Zhang, Preum, Stankovic, and Ma}]{chen2022cityspec}
Chen, Z.; Li, I.; Zhang, H.; Preum, S.; Stankovic, J.~A.; and Ma, M. 2022{\natexlab{a}}.
\newblock Cityspec: An intelligent assistant system for requirement specification in smart cities.
\newblock In \emph{2022 IEEE International Conference on Smart Computing (SMARTCOMP)}, 32--39. IEEE.

\bibitem[{Chen et~al.(2023)Chen, Li, Zhang, Preum, Stankovic, and Ma}]{chen2023cityspecshield}
Chen, Z.; Li, I.; Zhang, H.; Preum, S.; Stankovic, J.~A.; and Ma, M. 2023.
\newblock CitySpec with shield: A secure intelligent assistant for requirement formalization.
\newblock \emph{Pervasive and Mobile Computing}, 92: 101802.

\bibitem[{Chen et~al.(2022{\natexlab{b}})Chen, Li, Zhang, Preurn, Stankovic, and Ma}]{chen2022intelligent}
Chen, Z.; Li, I.; Zhang, H.; Preurn, S.; Stankovic, J.~A.; and Ma, M. 2022{\natexlab{b}}.
\newblock An Intelligent Assistant for Converting City Requirements to Formal Specification.
\newblock In \emph{2022 IEEE International Conference on Smart Computing (SMARTCOMP)}, 174--176. IEEE.

\bibitem[{Chipman et~al.(2022)Chipman, George, McCulloch, and Shively}]{chipman2022mbart}
Chipman, H.~A.; George, E.~I.; McCulloch, R.~E.; and Shively, T.~S. 2022.
\newblock mBART: multidimensional monotone BART.
\newblock \emph{Bayesian Analysis}, 17(2): 515--544.

\bibitem[{Damerau(1964)}]{damerau1964technique}
Damerau, F.~J. 1964.
\newblock A technique for computer detection and correction of spelling errors.
\newblock \emph{Communications of the ACM}, 7(3): 171--176.

\bibitem[{Devlin et~al.(2018)Devlin, Chang, Lee, and Toutanova}]{devlin2018bert}
Devlin, J.; Chang, M.-W.; Lee, K.; and Toutanova, K. 2018.
\newblock Bert: Pre-training of deep bidirectional transformers for language understanding.
\newblock \emph{arXiv preprint arXiv:1810.04805}.

\bibitem[{Diefenbach et~al.(2018)Diefenbach, Lopez, Singh, and Maret}]{diefenbach2018core}
Diefenbach, D.; Lopez, V.; Singh, K.; and Maret, P. 2018.
\newblock Core techniques of question answering systems over knowledge bases: a survey.
\newblock \emph{Knowledge and Information systems}, 55: 529--569.

\bibitem[{Face(2022)}]{huggingface_bert_qa_2022}
Face, H. 2022.
\newblock BERT: Pre-trained Transformer for Question Answering.
\newblock Accessed: 2023-08-03.

\bibitem[{Gal and Ghahramani(2016)}]{gal2016dropout}
Gal, Y.; and Ghahramani, Z. 2016.
\newblock Dropout as a bayesian approximation: Representing model uncertainty in deep learning.
\newblock In \emph{international conference on machine learning}, 1050--1059. PMLR.

\bibitem[{Hendrycks et~al.(2021)Hendrycks, Burns, Chen, and Ball}]{hendrycks2021cuad}
Hendrycks, D.; Burns, C.; Chen, A.; and Ball, S. 2021.
\newblock CUAD: An Expert-Annotated NLP Dataset for Legal Contract Review.
\newblock \emph{arXiv preprint arXiv:2103.06268}.

\bibitem[{Hochreiter and Schmidhuber(1997)}]{hochreiter1997long}
Hochreiter, S.; and Schmidhuber, J. 1997.
\newblock Long short-term memory.
\newblock \emph{Neural computation}, 9(8): 1735--1780.

\bibitem[{H{\"u}llermeier and Waegeman(2021)}]{hullermeier2021aleatoric}
H{\"u}llermeier, E.; and Waegeman, W. 2021.
\newblock Aleatoric and epistemic uncertainty in machine learning: An introduction to concepts and methods.
\newblock \emph{Machine Learning}, 110: 457--506.

\bibitem[{Joshi et~al.(2017)Joshi, Choi, Weld, and Zettlemoyer}]{joshi2017triviaqa}
Joshi, M.; Choi, E.; Weld, D.~S.; and Zettlemoyer, L. 2017.
\newblock Triviaqa: A large scale distantly supervised challenge dataset for reading comprehension.
\newblock \emph{arXiv preprint arXiv:1705.03551}.

\bibitem[{Kim(2014)}]{kim2014convolutional}
Kim, Y. 2014.
\newblock Convolutional neural networks for sentence classification.
\newblock \emph{arXiv preprint arXiv:1408.5882}.

\bibitem[{Kowsari et~al.(2019)Kowsari, Jafari~Meimandi, Heidarysafa, Mendu, Barnes, and Brown}]{kowsari2019text}
Kowsari, K.; Jafari~Meimandi, K.; Heidarysafa, M.; Mendu, S.; Barnes, L.; and Brown, D. 2019.
\newblock Text classification algorithms: A survey.
\newblock \emph{Information}, 10(4): 150.

\bibitem[{Lai et~al.(2015)Lai, Xu, Liu, and Zhao}]{lai2015recurrent}
Lai, S.; Xu, L.; Liu, K.; and Zhao, J. 2015.
\newblock Recurrent convolutional neural networks for text classification.
\newblock In \emph{Twenty-ninth AAAI conference on artificial intelligence}.

\bibitem[{Lin(2004)}]{lin-2004-rouge}
Lin, C.-Y. 2004.
\newblock {ROUGE}: A Package for Automatic Evaluation of Summaries.
\newblock In \emph{Text Summarization Branches Out}, 74--81. Barcelona, Spain: Association for Computational Linguistics.

\bibitem[{Ma et~al.(2021{\natexlab{a}})Ma, Bartocci, Lifland, Stankovic, and Feng}]{ma2021novel}
Ma, M.; Bartocci, E.; Lifland, E.; Stankovic, J.~A.; and Feng, L. 2021{\natexlab{a}}.
\newblock A novel spatial--temporal specification-based monitoring system for smart cities.
\newblock \emph{IEEE Internet of Things Journal}, 8(15): 11793--11806.

\bibitem[{Ma et~al.(2019)Ma, Preum, Ahmed, T{\"a}rneberg, Hendawi, and Stankovic}]{ma2019data}
Ma, M.; Preum, S.~M.; Ahmed, M.~Y.; T{\"a}rneberg, W.; Hendawi, A.; and Stankovic, J.~A. 2019.
\newblock Data sets, modeling, and decision making in smart cities: A survey.
\newblock \emph{ACM Transactions on Cyber-Physical Systems}, 4(2): 1--28.

\bibitem[{Ma et~al.(2021{\natexlab{b}})Ma, Stankovic, Bartocci, and Feng}]{ma2021predictive}
Ma, M.; Stankovic, J.; Bartocci, E.; and Feng, L. 2021{\natexlab{b}}.
\newblock Predictive monitoring with logic-calibrated uncertainty for cyber-physical systems.
\newblock \emph{ACM Transactions on Embedded Computing Systems (TECS)}, 20(5s): 1--25.

\bibitem[{Ma, Stankovic, and Feng(2018)}]{ma2018cityresolver}
Ma, M.; Stankovic, J.~A.; and Feng, L. 2018.
\newblock Cityresolver: a decision support system for conflict resolution in smart cities.
\newblock In \emph{2018 ACM/IEEE 9th International Conference on Cyber-Physical Systems (ICCPS)}, 55--64. IEEE.

\bibitem[{Ma, Stankovic, and Feng(2021)}]{ma2021toward}
Ma, M.; Stankovic, J.~A.; and Feng, L. 2021.
\newblock Toward formal methods for smart cities.
\newblock \emph{Computer}, 54(9): 39--48.

\bibitem[{Manoj and Baker(2007)}]{manoj2007communication}
Manoj, B.~S.; and Baker, A.~H. 2007.
\newblock Communication challenges in emergency response.
\newblock \emph{Communications of the ACM}, 50(3): 51--53.

\bibitem[{Minaee et~al.(2021)Minaee, Kalchbrenner, Cambria, Nikzad, Chenaghlu, and Gao}]{minaee2021deep}
Minaee, S.; Kalchbrenner, N.; Cambria, E.; Nikzad, N.; Chenaghlu, M.; and Gao, J. 2021.
\newblock Deep learning--based text classification: a comprehensive review.
\newblock \emph{ACM computing surveys (CSUR)}, 54(3): 1--40.

\bibitem[{Miro{\'n}czuk and Protasiewicz(2018)}]{mironczuk2018recent}
Miro{\'n}czuk, M.~M.; and Protasiewicz, J. 2018.
\newblock A recent overview of the state-of-the-art elements of text classification.
\newblock \emph{Expert Systems with Applications}, 106: 36--54.

\bibitem[{Mukhopadhyay et~al.(2022)Mukhopadhyay, Pettet, Vazirizade, Lu, Jaimes, El~Said, Baroud, Vorobeychik, Kochenderfer, and Dubey}]{mukhopadhyay2022review}
Mukhopadhyay, A.; Pettet, G.; Vazirizade, S.~M.; Lu, D.; Jaimes, A.; El~Said, S.; Baroud, H.; Vorobeychik, Y.; Kochenderfer, M.; and Dubey, A. 2022.
\newblock A review of incident prediction, resource allocation, and dispatch models for emergency management.
\newblock \emph{Accident Analysis \& Prevention}, 165: 106501.

\bibitem[{NYC911(2022)}]{nycnextgen911}
NYC911. 2022.
\newblock {Next-Gen 911 on Target for 2024 Completion}.
\newblock \url{https://www.nyc.gov/content/oti/pages/press-releases/next-gen-911-on-target-2024-completion}.
\newblock Accessed: 08-15-2023.

\bibitem[{OpenAI(2022)}]{openai_whisper_2022}
OpenAI. 2022.
\newblock {Whisper: Speech Recognition}.
\newblock Accessed: 2023-08-03.

\bibitem[{Ouyang et~al.(2022)Ouyang, Wu, Jiang, Almeida, Wainwright, Mishkin, Zhang, Agarwal, Slama, Ray et~al.}]{ouyang2022training}
Ouyang, L.; Wu, J.; Jiang, X.; Almeida, D.; Wainwright, C.; Mishkin, P.; Zhang, C.; Agarwal, S.; Slama, K.; Ray, A.; et~al. 2022.
\newblock Training language models to follow instructions with human feedback.
\newblock \emph{Advances in Neural Information Processing Systems}, 35: 27730--27744.

\bibitem[{Papineni et~al.(2002)Papineni, Roukos, Ward, and Zhu}]{papineni2002bleu}
Papineni, K.; Roukos, S.; Ward, T.; and Zhu, W.-J. 2002.
\newblock BLEU: a method for automatic evaluation of machine translation.
\newblock In \emph{Proceedings of the 40th annual meeting of the Association for Computational Linguistics}, 311--318.

\bibitem[{Poggi and Mattoccia(2016)}]{poggi2016learning}
Poggi, M.; and Mattoccia, S. 2016.
\newblock Learning from scratch a confidence measure.
\newblock In \emph{Bmvc}, volume~2, 4.

\bibitem[{Poggi, Tosi, and Mattoccia(2017)}]{poggi2017quantitative}
Poggi, M.; Tosi, F.; and Mattoccia, S. 2017.
\newblock Quantitative evaluation of confidence measures in a machine learning world.
\newblock In \emph{Proceedings of the IEEE International Conference on Computer Vision}, 5228--5237.

\bibitem[{Raffel et~al.(2019)Raffel, Shazeer, Roberts, Lee, Narang, Matena, Zhou, Li, and Liu}]{raffel2019t5}
Raffel, C.; Shazeer, N.; Roberts, A.; Lee, K.; Narang, S.; Matena, M.; Zhou, Y.; Li, W.; and Liu, P.~J. 2019.
\newblock Exploring the Limits of Transfer Learning with a Unified Text-to-Text Transformer.
\newblock In \emph{Advances in Neural Information Processing Systems (NeurIPS)}.

\bibitem[{Rajpurkar, Jia, and Liang(2018)}]{rajpurkar2018know}
Rajpurkar, P.; Jia, R.; and Liang, P. 2018.
\newblock Know what you don't know: Unanswerable questions for SQuAD.
\newblock \emph{arXiv preprint arXiv:1806.03822}.

\bibitem[{Rajpurkar et~al.(2016)Rajpurkar, Zhang, Lopyrev, and Liang}]{rajpurkar2016squad}
Rajpurkar, P.; Zhang, J.; Lopyrev, K.; and Liang, P. 2016.
\newblock Squad: 100,000+ questions for machine comprehension of text.
\newblock \emph{arXiv preprint arXiv:1606.05250}.

\bibitem[{Reimers and Gurevych(2019)}]{reimers-2019-sentence-bert}
Reimers, N.; and Gurevych, I. 2019.
\newblock Sentence-BERT: Sentence Embeddings using Siamese BERT-Networks.
\newblock \emph{arXiv preprint arXiv:1908.10084}.

\bibitem[{Sanh et~al.(2019)Sanh, Debut, Chaumond, and Wolf}]{sanh2019distilbert}
Sanh, V.; Debut, L.; Chaumond, J.; and Wolf, T. 2019.
\newblock DistilBERT, a distilled version of BERT: smaller, faster, cheaper and lighter.
\newblock \emph{arXiv preprint arXiv:1910.01108}.

\bibitem[{Saxon et~al.(2022)Saxon, Villena, Wilburn, Andersen, Maloney, and Jacobson}]{2021annual}
Saxon, N.; Villena, P.; Wilburn, S.; Andersen, S.; Maloney, D.; and Jacobson, R. 2022.
\newblock Annual Survey of Public Employment \& Payroll Summary Report: 2021.
\newblock \emph{US Census Bureau}.

\bibitem[{Sun, Bocchini, and Davison(2020)}]{sun2020applications}
Sun, W.; Bocchini, P.; and Davison, B.~D. 2020.
\newblock Applications of artificial intelligence for disaster management.
\newblock \emph{Natural Hazards}, 103(3): 2631--2689.

\bibitem[{{suno-ai}(2023)}]{bark}
{suno-ai}. 2023.
\newblock Text-Prompted Generative Audio Model.
\newblock \url{https://github.com/suno-ai/bark}.
\newblock Accessed: 9 August 2023.

\bibitem[{Vaswani et~al.(2017)Vaswani, Shazeer, Parmar, Uszkoreit, Jones, Gomez, Kaiser, and Polosukhin}]{vaswani2017attention}
Vaswani, A.; Shazeer, N.; Parmar, N.; Uszkoreit, J.; Jones, L.; Gomez, A.~N.; Kaiser, L.; and Polosukhin, I. 2017.
\newblock Attention is all you need.
\newblock \emph{Advances in neural information processing systems}, 30.

\bibitem[{Wex et~al.(2014)Wex, Schryen, Feuerriegel, and Neumann}]{wex2014emergency}
Wex, F.; Schryen, G.; Feuerriegel, S.; and Neumann, D. 2014.
\newblock Emergency response in natural disaster management: Allocation and scheduling of rescue units.
\newblock \emph{European Journal of Operational Research}, 235(3): 697--708.

\bibitem[{Wolf et~al.(2019)Wolf, Debut, Sanh, Chaumond, Delangue, Moi, Cistac, Rault, Louf, Funtowicz, and Brew}]{wolf2019huggingface}
Wolf, T.; Debut, L.; Sanh, V.; Chaumond, J.; Delangue, C.; Moi, A.; Cistac, P.; Rault, T.; Louf, R.; Funtowicz, M.; and Brew, J. 2019.
\newblock HuggingFace's Transformers: State-of-the-art Natural Language Processing.
\newblock \url{https://github.com/huggingface/transformers}.
\newblock Accessed: 2023-02-20.

\bibitem[{Zaheer et~al.(2020)Zaheer, Guruganesh, Dubey, Ainslie, Alberti, Ontanon, Pham, Ravula, Wang, Yang et~al.}]{zaheer2020big}
Zaheer, M.; Guruganesh, G.; Dubey, K.~A.; Ainslie, J.; Alberti, C.; Ontanon, S.; Pham, P.; Ravula, A.; Wang, Q.; Yang, L.; et~al. 2020.
\newblock Big bird: Transformers for longer sequences.
\newblock \emph{Advances in neural information processing systems}, 33: 17283--17297.

\end{thebibliography}
\appendix
\begin{center}
\section*{Appendix}
\end{center}

\section{Detailed Introduction on Dataset}

\begin{figure*}[htbp]
    \centering
    \includegraphics[width=0.9\textwidth]{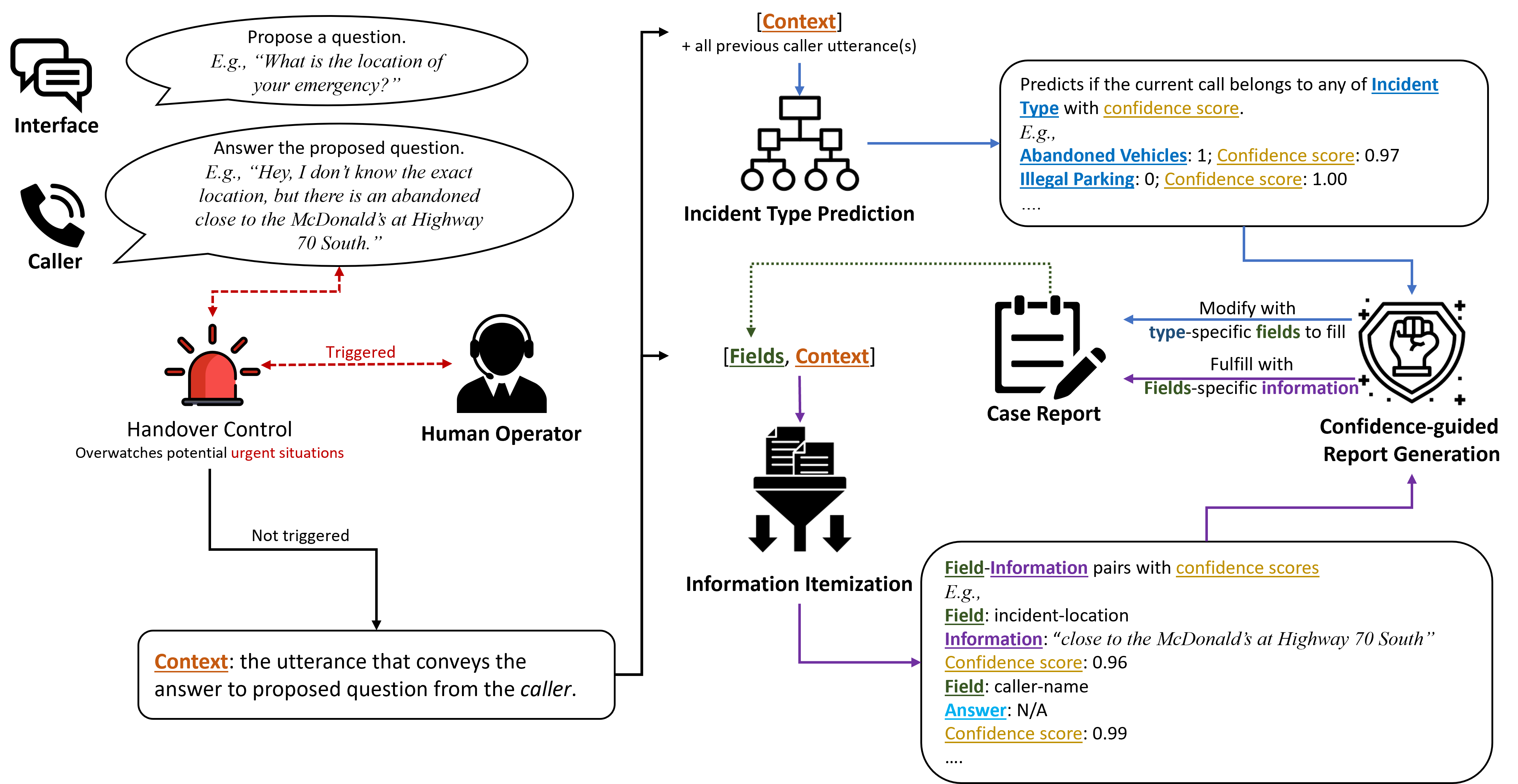}
    \caption{Running Example within One Turn of Conversation}
    \label{fig:running_example}
    \vspace{-0.5cm}
\end{figure*}

Our dataset includes real-world audio recordings covering from November 2022 to February 2023 across over 11 non-emergency types, we focus on the most common 11 incident types in this paper. Incident types are carefully annotated by dispatchers from the emergency response center.

In this section, we introduce our dataset in more detail. Initially, the dataset consists of 11,796 real recordings from the local call center. To convert the audio files in textual format for Auto311 to learn, we transcribe the audio with speaker diarization \cite{openai_whisper_2022, Bredin_pyannote_2020}. Following is one sample transcription from our dataset reporting an abandoned vehicle with private and sensitive information masked.

\noindent[00:00:13.727]--[00:00:16.458] $\mathsf{Dispatcher}$: Police Fire and Medical. 

\noindent[00:00:17.121]--[00:00:36.458] $\mathsf{Caller}$: Oh, good morning. Um, there is a car that seems to be abandoned, uh, across the street. And I just wondered if someone could check it out. It hasn't moved in weeks and weeks and weeks, and it's got kind of tinted windows and it's just creepy.

\noindent[00:00:36.863]--[00:00:38.061] $\mathsf{Dispatcher}$:  Okay, where is it located?

\noindent[00:00:39.023]--[00:01:05.483] $\mathsf{Caller}$: It is across the street from \textit{\#masked\_address}, and that's \textit{\#masked\_address}. It's in front of like a very small green space. There are kids that play in that green space, and we just all kind of, you know, creeped out about it.

\noindent[00:01:05.685]--[00:01:11.372] $\mathsf{Dispatcher}$:  Okay, so \textit{\#masked\_address}. What kind of vehicle is it?

\noindent[00:01:12.182]--[00:01:31.842] $\mathsf{Caller}$:  It is a very dark grey Volvo I believe is what it is. It's a very dark grey car and the windows are tinted just a bit.

\noindent[00:01:35.740]--[00:01:47.316] $\mathsf{Dispatcher}$: Alright, and what's your first and last name?

\noindent[00:01:48.261]--[00:01:49.746] $\mathsf{Caller}$: Okay, my name is \textit{\#masked\_personal\_information}.

\noindent[00:01:50.421]--[00:01:52.598] $\mathsf{Dispatcher}$: Did you want to speak to an officer when they come out?

\noindent[00:01:53.340]--[00:01:55.517] $\mathsf{Caller}$:  No, there's no need for that.

\noindent[00:01:55.956]--[00:02:05.204] $\mathsf{Dispatcher}$:  Alright, I will get someone out there across \textit{\#masked\_address} as soon as we can, okay?

\noindent[00:02:05.912]--[00:02:09.642] $\mathsf{Caller}$:  That sounds great. Thank you so much and I hope you have a real good day.

\noindent[00:02:09.574]--[00:02:10.013] $\mathsf{Dispatcher}$:  You too, bye.

\section{One Example Turn of Auto311}
A further example in Auto311 is also provided in Figure \ref{fig:running_example}.

\section{Handover Control Patterns}

We denote the ongoing user utterance as $\mathsf{S}$, the set of patterns as $p_0, p_1, ..., p_n \subset \mathsf{P}$, and the whole process as a boolean function $\mathsf{is\_trigger(S, {P})}$. Our rule-based procedure can be concluded using the following pseudo-logic: 

``if any pattern $p$ of $\mathsf{P}$ exists in $\mathsf{S}$, then $\mathsf{is\_trigger(S, {P})}$ returns true, the handover control is triggered, and the system interactions will be ceased and the call will be rerouted to a real operator immediately. Otherwise, $\mathsf{is\_trigger(S, {P})}$ returns false and the handover control continues over-watching the ongoing call.'' 

\begin{table}[h]
\centering
\scriptsize
\caption{Example patterns in handover control}
\begin{tabular}{||c|c|c||}
\hline
Cases                                       & Example Patterns             & Example Texts      \\ \hline\hline
\multirow{2}{*}{Request Human Operators} & {[}NP$\ast${]}                    & real human         \\ \cline{2-3} 
                                            & {[}VP$\ast${]}                    & end the call       \\ \hline
\multirow{2}{*}{Alert Potential Urgency} & {[}PRP{]}{[}be{]}{[}ADJP$\ast${]} & he is unresponsive \\ \cline{2-3} 
                                            & {[}VP$\ast${]}{[}be{]}{[}NP$\ast${]}   & guns are fired     \\ \hline
\end{tabular}
\label{tab:patterns}
\end{table}

Here we provide several patterns in Table \ref{tab:patterns}, where VP, NP, PRP, ADJP, and PP refer to verb phrases, noun phrases, personal phrases, adjective phrases, and prepositional phrases. We mark phrases that need to be maintained during runtime using sensitive keywords using stars ($\ast$). Note, this is not an exhaustive table, both patterns and sensitive keywords will be extended during usage.

\section{Phone tree of Incident Types}
\begin{figure*}[ht]
\centering
\rotatebox{90}{
  \includegraphics[width=1.4\textwidth]{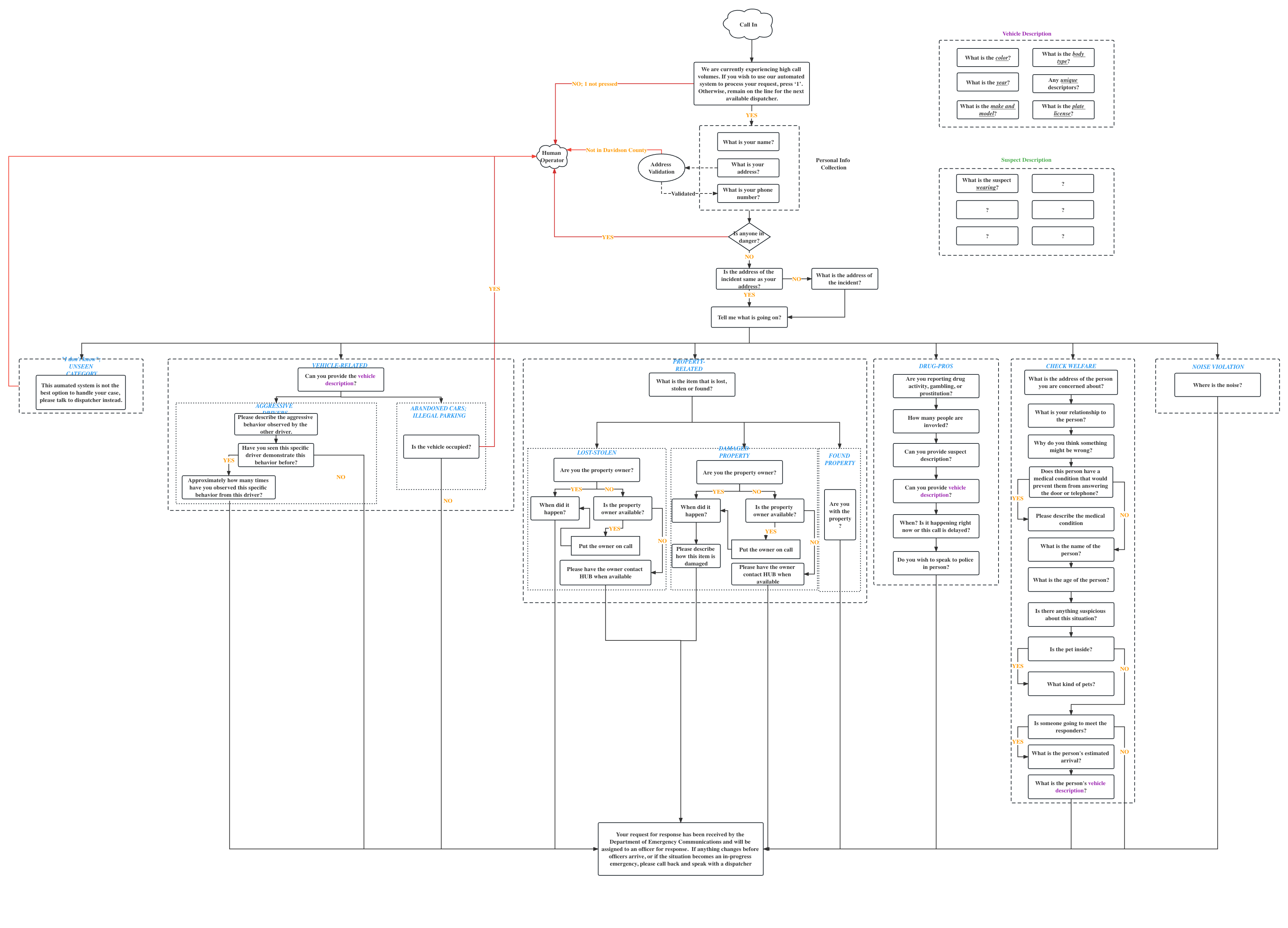} 
}
\caption{Phone Tree for Incident Handling} 
\label{fig:phone_tree}
\end{figure*}

We work with city authorities and create the following phone tree for different incident types, including 11 types of incidents including, illegal parking, abandoned vehicle, aggressive driver, lost-stolen, damaged property, found property, drug pros, check welfare, and noise violation. Each event has its only specific fields to fill alongside each call. The phone tree is in Figure \ref{fig:phone_tree}, we only show 9 incident types here due to the page size.

\subsection{Additional Evaluation} 

\begin{table}[]
\footnotesize
\begin{tabular}{||c|cccc||}
\hline
          & \multicolumn{4}{c||}{Illegal Parking(binary)}                                                                                                 \\ \hline\hline
Metric    & \multicolumn{1}{c|}{Precision}         & \multicolumn{1}{c|}{Recall}            & \multicolumn{1}{c|}{F-1}               & Accuracy          \\ \hline
LSTM      & \multicolumn{1}{c|}{53.85\%}           & \multicolumn{1}{c|}{87.79\%}           & \multicolumn{1}{c|}{70.00\%}           & 53.85\%           \\ \hline
CNN       & \multicolumn{1}{c|}{0.00\%}                 & \multicolumn{1}{c|}{0.00\%}                 & \multicolumn{1}{c|}{0.00\%}                 & 0.00\%                 \\ \hline
RCNN      & \multicolumn{1}{c|}{0.00\%}                 & \multicolumn{1}{c|}{0.00\%}                 & \multicolumn{1}{c|}{0.00\%}                 & 0.00\%                 \\ \hline
RNN       & \multicolumn{1}{c|}{25.00\%}           & \multicolumn{1}{c|}{28.57\%}           & \multicolumn{1}{c|}{26.67\%}           & 15.38\%           \\ \hline
Self-Attn & \multicolumn{1}{c|}{0.00\%}                 & \multicolumn{1}{c|}{0.00\%}                 & \multicolumn{1}{c|}{0.00\%}                 & 0.00\%                 \\ \hline
Attention & \multicolumn{1}{c|}{55.56\%}           & \multicolumn{1}{c|}{71.42\%}           & \multicolumn{1}{c|}{62.50\%}           & 53.85\%           \\ \hline
Bert      & \multicolumn{1}{c|}{100.00\%} & \multicolumn{1}{c|}{100.00\%} & \multicolumn{1}{c|}{100.00\%} & 100.00\% \\ \hline
\textbf{Auto311}   & \multicolumn{1}{c|}{\textbf{100.00\%}} & \multicolumn{1}{c|}{\textbf{100.00\%}} & \multicolumn{1}{c|}{\textbf{100.00\%}} & \textbf{100.00\%} \\ \hline
\end{tabular}
\label{table:rest_eval}
\caption{Auto311 on Incident Type Prediction}
\end{table}

\subsubsection{Evaluation on Incident Type Prediction}
Here we provide the rest evaluation of Auto311 on incident prediction, specifically on illegal parking and found property. The last layer of the incident type prediction module contains a binary classification of illegal parking and found property, see Table 4 for more details. As we can tell from the table, with the introduction of confidence guidance, Auto311 leverages the strong prior knowledge from BERT, yielding 100.00\% F-1 scores on both incident types.


\subsubsection{Validating Proposed Metric for Text Comparison}

We conduct an evaluation to assess the effectiveness of our proposed text comparison metric, targeting the question ``How does this new text comparison work in this specific scenario?''. For this evaluation, we manually selected three distinct groups of text pairs:
\begin{itemize}
    \item \textbf{Group one} consists of text pairs that are entirely dissimilar, for instance, ``65 South exit 92'' and ``Silver Camaro.'' These pairs serve as a benchmark to evaluate how well the metric can discern vastly different information.
    \item \textbf{Group two} comprises pairs that exhibit slight differences in their content, but these discrepancies are not significant enough to significantly impact the dispatcher's decision-making process. For example, pairs like ``an SUV type truck'' and ``It's like an SUV type truck, maybe a Tahoe'' fall under this category.
    \item \textbf{Group three} contains pairs with identical or highly similar information, which are relevant for dispatchers to complete internal reports. Examples of such pairs include ``on the West End Ave'' and ``West End Ave.''
\end{itemize}

To test the consistency scores, we utilize traditional metrics like BLEU \cite{papineni2002bleu}, Damerau–Levenshtein Distance (DLD) \cite{damerau1964technique}, and ROUGE \cite{lin-2004-rouge}, in addition to our modified metric. The evaluation aims to compare how each metric performs in distinguishing differences and similarities within the text pairs across these distinct groups. This analysis will provide valuable insights into the efficacy of our proposed metric compared to established text comparison metrics.

\begin{figure}[h]
    \centering
    \includegraphics[width=0.40\textwidth]{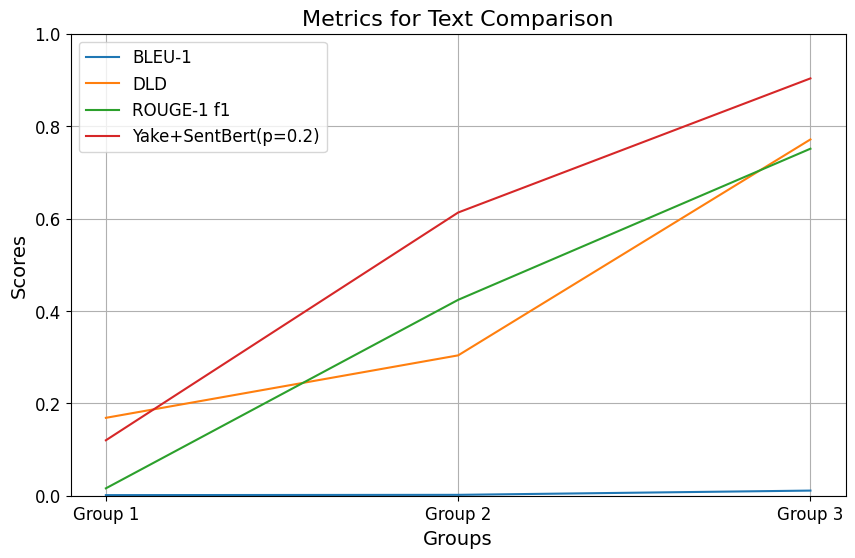}
    \caption{Metric for Text Comparison}
    \label{fig:textcompare}
    \vspace{-0.5cm}
\end{figure}

Analysis of the plot reveals certain transitional metrics display an upward trend while failing to furnish a fair assessment. The ideal metric yields low scores for group one pair comparisons and high scores for group three pairs. As Figure \ref{fig:textcompare} displays, although exhibiting increased scores from group one to group three, DLD and ROUGE do not account for the robust correlations among text pairs in group three.


\textit{Ultimately, from the results, it indicates our proposed text comparison metric proves more effective in assessing texts for consistency in this non-emergency dispatching scenario.}


\section{Discussion and Future Work}
This section discusses potential future work to enhance our proposed system based on additional findings from dataset review and development planning.

\subsubsection{Address Validation}
Location information is critical in emergency response systems. However, callers often describe locations using landmarks, not addresses. Descriptions like ``the McDonald's at Charlotte Pike'' are implicit. Translating these to explicit addresses or coordinates could better locate incidents. This work highlights address information from caller utterances without additional validation.


\subsubsection{Redundant Call and Callback Handling}
Redundant calls frequently occur reporting identical incidents. For instance, when witnesses spot a highway car crash, multiple people may call to report the same event, providing descriptions like ``there is a car wreck on I-440, milestone 76'' or ``there is a severe car crash on Interstate 440.'' While calling about the same incident, each call can still furnish unique unobtainable information. Similarly, caller callbacks regarding the same incident often provide previously omitted details, e.g., ``Hey, I just called a few minutes ago reporting an aggressive driver on I-40, I think the driver is in a blue Toyota, I just saw him.'' Hence, skipping or terminating ostensibly redundant calls proves inappropriate. Instead, solutions should strategically emphasize novel information despite referring to a previously reported incident. Presently, we treat each incoming call equivalently without assigning differential importance to any information we attempt to collect.

\begin{table}[]
\footnotesize
\begin{tabular}{||c|c|c|c|c||}
\hline
          & Minor Crash Report & Lost-Stolen & Aggressive Drivers & Damaged Property \\ \hline\hline
LSTM      & 56.47\%            & 0.00\%      & 53.85\%            & 0.00\%           \\ \hline
CNN       & 75.86\%            & 85.71\%     & 72.72\%            & 40.00\%          \\ \hline
RCNN      & 90.57\%            & 82.35\%     & 61.54\%            & 76.92\%          \\ \hline
RNN       & 63.33\%            & 40.00\%     & 52.71\%            & 44.44\%          \\ \hline
Self-Attn & 88.46\%            & 88.89\%     & 66.67\%            & 66.67\%          \\ \hline
Attention & 91.69\%            & 62.50\%     & 50.00\%            & 54.55\%          \\ \hline
Bert      & 95.04\%            & 95.60\%     & 92.31\%            & 88.89\%          \\ \hline
\textbf{Auto311}   & \textbf{95.71\%}            & \textbf{96.70\%}     & \textbf{93.75\%}            & \textbf{94.12\%}          \\ \hline
\end{tabular}
\end{table}
\end{document}